\pdfoutput=1

\documentclass[11pt]{article}

\usepackage{acl}
\usepackage{times}
\usepackage{latexsym}

\usepackage[T1]{fontenc}

\usepackage[utf8]{inputenc}

\usepackage{microtype}

\usepackage{multirow}
\usepackage{subfigure}
\usepackage{bm}
\usepackage{amssymb}
\usepackage{verbatim}
\usepackage{graphicx}
\usepackage{amsmath}
\usepackage{amsthm}
\usepackage{booktabs}
\usepackage[noend]{algpseudocode}
\usepackage[ruled,vlined,linesnumbered]{algorithm2e} 
\usepackage{stfloats}

\title{Classical Sequence Match is a Competitive Few-Shot One-Class Learner}

\author{Mengting Hu\textsuperscript{1} \quad Hang Gao\textsuperscript{2}\thanks{\; Hang Gao is the corresponding author.} \quad Yinhao Bai\textsuperscript{1} \quad {\bf Mingming Liu\textsuperscript{1}} \\
\textsuperscript{1} College of Software, Nankai University \\
\textsuperscript{2} Institute for Public Safety Research, Tsinghua University \\
{\tt mthu@nankai.edu.cn,} {\tt gaohang@mail.tsinghua.edu.cn} \\ {\tt yinhao@mail.nankai.edu.cn,}
{\tt liumingming@nankai.edu.cn}
}

\begin{document}
\maketitle
\begin{abstract}
Nowadays, transformer-based models gradually become the \emph{``default choice''} for artificial intelligence pioneers. The models also show superiority even in the few-shot scenarios. In this paper, we revisit the classical methods and propose a new few-shot alternative. Specifically, we investigate the few-shot one-class problem, which actually takes a known sample as a reference to detect whether an unknown instance belongs to the same class. This problem can be studied from the perspective of sequence match. It is shown that with meta-learning, the classical sequence match method, i.e. Compare-Aggregate, significantly outperforms transformer ones. The classical approach requires much less training cost. Furthermore, we perform an empirical comparison between two kinds of sequence match approaches under simple fine-tuning and meta-learning. Meta-learning causes the transformer models' features to have high-correlation dimensions. The reason is closely related to the number of layers and heads of transformer models. Experimental codes and data are available at \url{https://github.com/hmt2014/FewOne}.
\end{abstract}

\section{Introduction}
When the labeled data is scarce in practical application, it is struggled to learn a well-performed model using deep learning algorithms. Yet annotating data costs much labor and time. Few-shot learning (FSL) intuitively addresses this obstacle \cite{koch2015siamese,vinyals2016matching,snell2017prototypical,finn2017model,sung2018learning}. FSL learns at the meta-task level, where each meta-task is formulated as inferring queries with the help of a support set \cite{vinyals2016matching}. Multiple meta-tasks facilitate the task-agnostic transferrable knowledge. Thus it can learn new knowledge fast after being taught only a few samples. Despite FSL has been well-studied, its one-class scenario \cite{frikha2020few} is less investigated. 

\begin{figure}[t]
\centering
\includegraphics[width=0.48\textwidth]{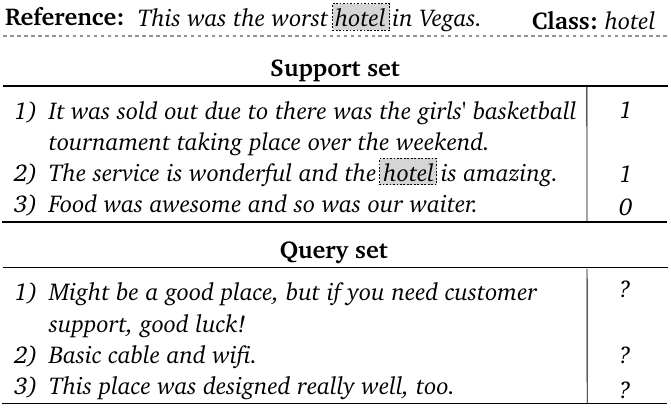}
\caption{Meta-task example in few-shot one-class text classification, where 1 denotes a positive instance and 0 denotes a negative one.}
\label{example}
\end{figure}

In this paper, following the one-class trait, we design each meta-task as a binary classification. It consists of a reference instance, a support set, and a query set (see Figure \ref{example}). The reference instance is one known sample of a class, which is exploited to tell whether an instance out of the support/query set belongs to the same class. Such purpose is consistent with sequence match, which also makes a decision for two sequences. Previous sequence match can mainly be categorized into two promising directions: \emph{classical methods}, e.g. Siamese Network \cite{koch2015siamese}, Compare-Aggregate (CA) \cite{wang2016compare}, and \emph{transformer-based method}, e.g. DistilBert \cite{sanh2019distilbert}, BERT \cite{devlin-etal-2019-bert}. 

In recent years, transformer models have already beaten classical ones in a wide range of tasks \cite{devlin-etal-2019-bert}. We wonder how two kinds of models perform under the few-shot one-class scenario. Consequently, it is presented that with meta-learning, classical sequence match method can significantly outperform transformer-based models. The classical models require much less training cost. Specifically, model-agnostic meta-learning (MAML) algorithm \cite{finn2017model} is a subtle bi-level optimizing approach that aims to learn a good parameter initialization. By introducing this algorithm, classical methods act as simple but competitive few-shot one-class learners. 

Furthermore, we make an empirical comparison between classical and transformer-based models under simple fine-tuning and meta-learning. Firstly, it is found that MAML has a more positive impact on the both sequence match approaches than simple fine-tuning. This suggests that a good parameter initialization is important for both of them. Secondly, MAML tends to make transformer models extract features with high-correlation dimensions. The bi-level optimization might cause the feature extraction layers of large models less trained. Yet the last classifying layer tends to be better learned relatively. We demonstrate that the high correlation is related to the number of heads and layers in the transformer.

In summary, our main contributions are as follows: (1) We present a simple but competitive few-shot one-class learner, which is based on the classical sequence match approach and meta-learning. Extensive experimental results show that this learner achieves significant improvements compared with transformer-based models. This provides new insights in the transformer-dominant era. (2) Based on the testbed provided by the above approaches, an empirical study is made to further reveal their underlining natures. New observations and conclusions are derived.

\section{Related Works}
\subsection{Few-Shot Learning}
Few-shot learning (FSL) \cite{fei2006one} deals with the practical problem of data scarcity in an intuitive way. It learns new knowledge fast with limited supervised information. An early work \cite{koch2015siamese} learns to detect whether two instances belong to the same class. Later, matching network \cite{vinyals2016matching} proposes to construct multiple meta-tasks in both the training and testing procedures. This setting becomes mainstream in the subsequent works, to name a few, distance-based methods \cite{snell2017prototypical,sung2018learning,garcia2017few,bao2019few}, optimization-based methods \cite{finn2017model,munkhdalai2017meta} or hallucination-based methods \cite{wang2018low,li2020adversarial}. Among them, MAML \cite{finn2017model} is special for ``model-agnostic'', indicating that this algorithm can be applied in any model. Therefore, we choose to further study its effects. To the best of our knowledge, it is seldom studied in transformer-based models. We provide an interesting empirical analysis in the experiments.

Recently, prompt-based fine-tuning \cite{gao2021making} also become popular in FSL. It classifies a template-based instance through the masked language model. This prediction manner bridges the gap between pre-training and fine-tuning. Its effectiveness in the few-shot one-class scenario is under-explored.

\subsection{One-Class Few-Shot Learning}
Recently some works discuss the one-class problem in FSL. Cumulative LEARning (CLEAR) \cite{kozerawski2018clear} uses transfer learning to model the decision boundary of SVM. One-way proto (OWP) \cite{kruspe2019one} is based on the prototypical network \cite{snell2017prototypical}. OWP computes the positive prototype by simply averaging the representations of instances. It designs a 0-vector as the negative prototype. The Euclidean distance with prototypes in the embedding space indicates that an instance is positive or negative. One-class MAML \cite{frikha2020few} proposes a simple data sampling strategy to ensure that the class-imbalance rate of the inner-level matches the test task. Different from them, we leverage the unique direction in natural language processing, i.e. sequence match, to study the one-class FSL.

\subsection{Sequence Match} 
Sequence match aims to make a decision for two sequences. Many tasks require to match sequences, such as text entailment \cite{bowman2015large}, machine comprehension \cite{tapaswi2016movieqa}, recommendation \cite{kraus2019personalized}, etc. A straightforward approach is to encode each sequence as a vector and then compare the two vectors to make a decision \cite{bowman2015large,feng2015applying}. However, a single vector is insufficient to match the important information between two sequences. Thus attention mechanism is adopted in this task \cite{rocktaschel2015reasoning}. 

Later, the Compare-Aggregate framework is proposed \cite{wang2016compare} for matching sequences, which has been widely studied. Its extended version usually considers the bidirectional information of two inputs \cite{bian2017compare,yoon2019compare}. One previous work \cite{ye2019multi} shows that matching and aggregation are effective in few-shot relation classification. We explore this framework in the few-shot one-class problem. More recently, pre-trained language models, e.g. BERT \cite{sanh2019distilbert} gain remarkable achievements in many sequence match tasks \cite{wang2020multi}.

\section{Methods}
In this work, two kinds of sequence match methods, including classical and transformer-based ones, are mainly investigated. Compare-Aggregate \cite{wang2016compare} is a promising classical method. We choose to study its extended version, i.e. Bidirectional Compare-Aggregate (BiCA) \cite{bian2017compare,yoon2019compare}, introduced in \S\ref{sub:bica}. The transformer-based sequence match \cite{sanh2019distilbert} is also presented in \S\ref{sub:transformer} briefly.

\subsection{Problem Definition}
Assume the training data $\mathcal{D}_{train}$ is composed of a set of training classes $\mathcal{C}_{train}$, and the testing data $\mathcal{D}_{test}$ has a set of classes $\mathcal{C}_{test}$, there are no overlapping between two class sets $\mathcal{C}_{train} \cap {\mathcal{C}_{test}}=\varnothing$. During training, we randomly sample a bunch of meta-tasks from $\mathcal{C}_{train}$. A meta-task is made as a binary classifier to detect one class, which is formulated as below. 

A meta-task contains a reference sentence $r$, a support set $S$ and a query set $Q$.
\begin{equation}
    \begin{split}
        S&=\{(x_s^1,y_s^1),(x_s^2,y_s^2),...(x_s^{|S|},y_s^{|S|})\} \\
        Q&=\{(x_q^1,y_q^1),(x_q^2,y_q^2),...(x_q^{|Q|},y_q^{|Q|})\}
    \end{split}
\end{equation}
where $x_s$/$x_q$ is an instance and $y_s$/$y_q$ denotes whether this instance belongs to the same class as the reference. $|S|$ and $|Q|$ indicate the number of instances in two sets, respectively. Many meta-tasks enable the model to extract task-agnostic knowledge, which is beneficial to the meta-tasks from the testing classes $\mathcal{C}_{test}$.

\begin{figure}[t]
\centering
\includegraphics[width=0.45\textwidth]{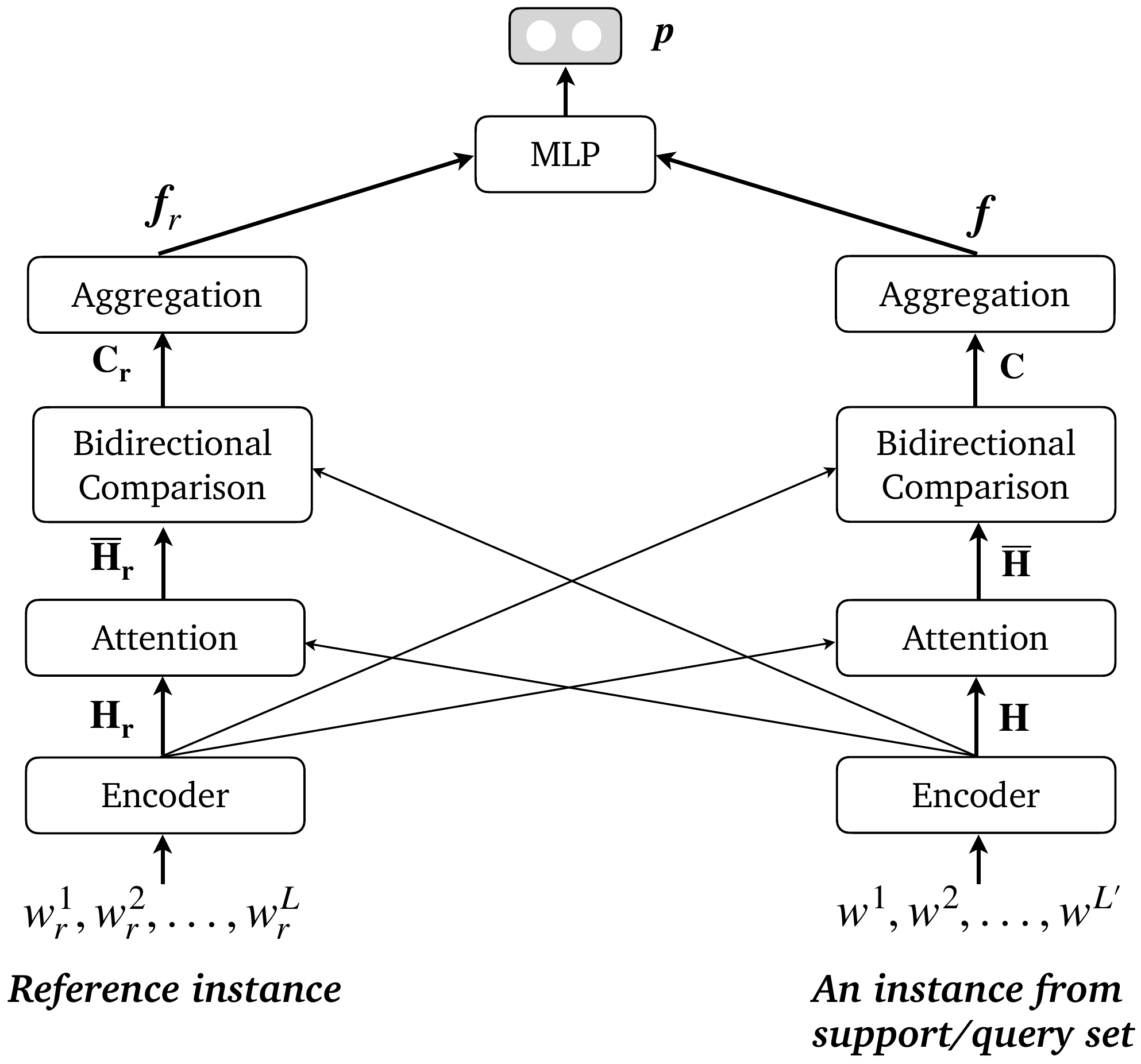}
\caption{The network architecture of BiCA. The parameters are shared by two input instances.}
\label{fig:bica}
\end{figure}

\begin{figure*}[t]
\centering
\includegraphics[width=0.9\textwidth]{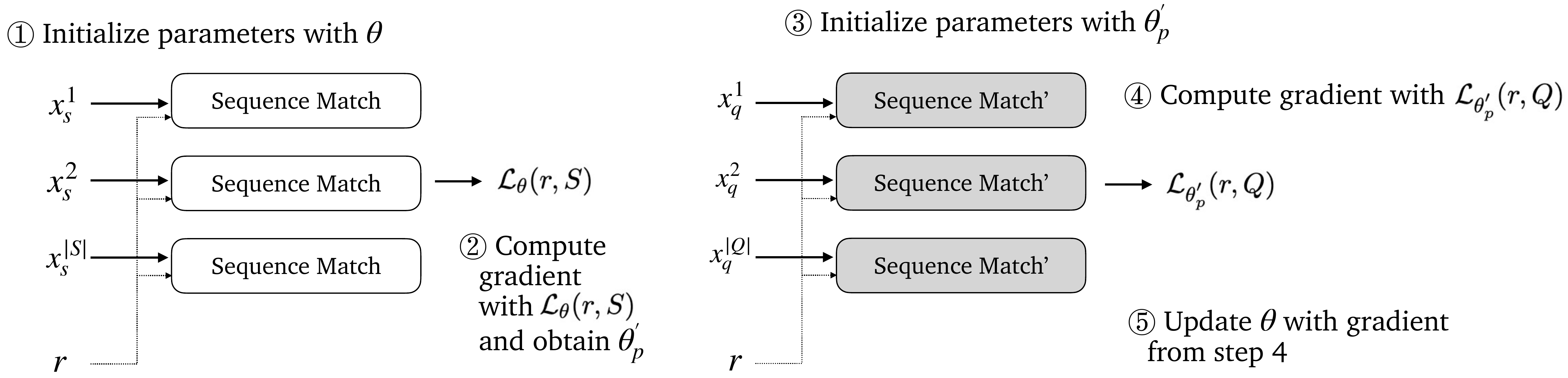}
\caption{Introducing meta-learning to sequence match approaches. The left part and the right part share the same architecture but employ different parameters in the meta-learning. The five steps are corresponding to the Algorithm \ref{alg_optimizing}. After training, the optimal parameter initialization $\theta$ for the testing classes is obtained.}
\label{fig:sequence_match}
\end{figure*}

\subsection{Classical Sequence Match}
\label{sub:bica}
In this section, we will introduce the components of Bidirectional Compare-Aggregate (BiCA) in detail.

\vspace{3pt}
\noindent
\textbf{Encoder} \; Given an input sentence with $L$ words, denoted as $\{w^1,w^2,...,w^L\}$, it is first mapped into an embedding sequence $\mathrm{\bf{E}}=\{\bm{e^1},\bm{e^2},...,\bm{e^L}\}$ by looking up the pre-trained GloVe embeddings \cite{pennington2014glove}. Then the embedding sequence is processed by the gate mechanism \cite{wang2016compare} to obtain contextualized information. This gating mechanism aims at remembering the meaningful words and filtering the less-important words in a sentence. 
\begin{equation}
    \mathrm{\bf{H}} = \mathrm{\sigma}(\mathrm{\bf{W^iE}}+\bm{b^i}){\odot}\mathrm{tanh}(\mathrm{\bf{W^uE}}+\bm{b^u})
\end{equation}
where $\mathrm{\bf{W^i}}$ and $\mathrm{\bf{W^u}}$ are parameter matrix, $\bm{b^i}$ and $\bm{b^u}$ are biases, $\odot$ is element-wise multiplication.

\vspace{3pt}
\noindent
\textbf{Attention} \; As depicted in Figure \ref{fig:bica}, the reference and an instance from support/query set are fed into the encoder, obtaining $\mathrm{\bf{H_r}}$ and $\mathrm{\bf{H}}$. Then the interaction between two inputs is computed through an attention mechanism.
\begin{equation}
\begin{split}
    \mathrm{\bf{\overline{H}_r}} &= \mathrm{\bf{H_r}} \cdot \mathrm{softmax}(\mathrm{\bf{H_r}}^{\mathrm{T}}\mathrm{\bf{H}}) \\  
    \mathrm{\bf{\overline{H}}} &= \mathrm{\bf{H}} \cdot \mathrm{softmax}(\mathrm{\bf{H}}^{\mathrm{T}}\mathrm{\bf{H_r}}) \\
\end{split}
\end{equation}

This attention mechanism is non-parametric since it only depends on the encoded representations $\mathrm{\bf{H_r}}$ and $\mathrm{\bf{H}}$. Such design reduces the reliance on parameters and focuses on learning the relationships between data. Hence, this helps better adapt to unseen classes.

\vspace{3pt}
\noindent
\textbf{Bidirectional Comparison} \; To compare the two instances, we adopt a simple word-level comparison function, i.e., element-wise multiplication $\odot$. 
\begin{equation}
    \mathrm{\bf{C_r}} = \mathrm{\bf{\overline{H}_r}} \odot \mathrm{\bf{H}} \;\;\;\;\;
    \mathrm{\bf{C}} = \mathrm{\bf{\overline{H}}} \odot \mathrm{\bf{H_r}} 
\end{equation}

The comparison function is also non-parametric for the purpose of adaptation. As shown in Figure \ref{fig:bica}, the encoded representations are applied in both the attention and bidirectional comparison modules to promote the mutual interaction of two inputs.

\vspace{3pt}
\noindent
\textbf{Aggregation} \; Following the original work \cite{wang2016compare}, the comparison representations are aggregated by convolution neural network (CNN) \cite{kim2014convolutional}. The convolution kernel slides over the comparison sequence to extract $n$-gram features, which tends to be helpful in matching.
\begin{equation}
    \bm{f_r} = \mathrm{CNN}(\mathrm{\bf{C_r}}) \;\;\;\;\;
    \bm{f} = \mathrm{CNN}(\mathrm{\bf{C}})
\end{equation}
where $\mathrm{CNN}(\cdot)$ is a convolution operation followed by max-pooling. Then the matching score is computed with the aggregation representations of two input sentences.
\begin{equation}
    \bm{p} = \mathrm{MLP}([\bm{f_r}, \bm{f}])
\end{equation}
where $\mathrm{MLP}$ is a single linear layer. $\bm{p}$ is a two-dimensional logits output.

\vspace{3pt}
\noindent
\textbf{Loss} \; The training objective of BiCA is the cross entropy loss.
\begin{equation}
    \mathcal{L}_{\theta}(r,S) = -\frac{1}{|S|}\sum_{|S|}\mathrm{log}P(y|\bm{p})
\label{equation_loss}
\end{equation}
where $y$ is the ground truth. $\theta$ represents all the parameters in the sequence match model.

\subsection{Transformer-Based Sequence Match}
\label{sub:transformer}
Transformer-based models, e.g. BERT \cite{devlin-etal-2019-bert} are pre-trained on a large-scale corpus, serving as foundational backbones for a wide range of natural language processing tasks. When aiming at sequence match, BERT utilizes two special tokens $\mathtt{[CLS]}$ and $\mathtt{[SEP]}$ to concatenate two sequences as a whole. For the two input instances shown in Figure \ref{fig:bica}, they are combined into $\{\mathtt{[CLS]},w_r^1,w_r^2,...,w_r^L,\mathtt{[SEP]},w^1,w^2,...,w^{L'}\}$, where the output of $\mathtt{[CLS]}$ is usually for classification and $\mathtt{[SEP]}$ is for separating two inputs. This combined sequence is then fed into BERT. The self-attention mechanism \cite{vaswani2017attention} in the transformer will compute the interaction between two inputs. Finally, the output of the first token $\mathtt{[CLS]}$ is adopted for inference. The training objective is also computed by Eq. (\ref{equation_loss}).

\begin{algorithm}[t]
\DontPrintSemicolon
\caption{Meta-Learning for Few-Shot One-Class Problem}
\label{alg_optimizing}
\KwIn{Training data from $\mathcal{D}_{train}$}
Randomly initialize $\theta$ \\
\Repeat{\emph{performance on the validation data set does not improve in 3 epochs.}}
{
{\bf Sample} positive classes $\mathcal{C}_p$ and negative classes $\mathcal{C}_n$ from $\mathcal{D}_{train}$
\; 
$\mathcal{C}_p\cap{\mathcal{C}_n}=\varnothing$  \;
\For{all $\mathcal{C}_p$}{
Construct a meta-task \;
Evaluate $\nabla_\theta{\mathcal{L}_\theta}(r,S)$ in Eq. (\ref{equation_loss}) \;
Compute adapted parameters with gradient descent: $\theta_p^{'}=\theta-\alpha\nabla_\theta{\mathcal{L}_\theta(r,S)}$ \;
}
Update $\theta\leftarrow\theta-\beta\nabla_\theta\sum_{\mathcal{C}_p}{\mathcal{L}_{\theta^{'}_p}(r,Q)}$
}
\end{algorithm}

\subsection{Meta-Learning for Sequence Match} 
In the few-shot one-class paradigm, a meta-task from the unseen class has a few labeled instances as the support set. To better leverage such knowledge, we introduce meta-learning to sequence match models, which is displayed in Figure \ref{fig:sequence_match} and Algorithm \ref{alg_optimizing}. Specifically, model-agnostic meta-learning (MAML) algorithm \cite{finn2017model} is chosen to investigate its impact on the sequence match approaches. This algorithm learns a good initialization of model parameters by maximizing the sensitivity of the loss function when adapting to new tasks \cite{song2020learning}.

\noindent
\textbf{Construct a Meta-Task} \; In Algorithm \ref{alg_optimizing} (line 6), given a positive class, we first sample $N+1$ instances from this class, which constitutes a reference instance $r$ and $N$ positive ones. Moreover, $N$ negative examples are also sampled from the negative classes $\mathcal{C}_n$. These positive and negative examples are mixed up randomly, which are further divided into the support set and query set.

Meta-learning trains in a bi-level way (see Algorithm \ref{alg_optimizing}), including the inner-level (line 8) and the outer-level (line 9) for the support set $S$ and the query set $Q$, respectively. This way will cause the gradients for updating parameters to propagate through more layers (line 9), i.e. twice as many as the number of network layers in a sequence match model. Its special effects on the transformer models are further discussed in \S\ref{subsec:discussion}. When evaluating, a model is initialized from the parameters $\theta$ trained by MAML on the training classes, which is then optimized with the support set on the testing classes. 

\begin{table}[t!]
\small
\begin{center}
\setlength{\tabcolsep}{1.8mm}{
\begin{tabular} {lc|ccc}
\toprule
     & & \textbf{Train} & \textbf{Validation} & \textbf{Test} \\
    \midrule
    Class Num & & 64 & 16 & 20  \\
    \midrule
    \multirow{2}{*}{Data Num} & single & 13677 & 3394 & 4671 \\
    & multi & 26643 & 6686 & 7929 \\
    \midrule
    \multirow{2}{*}{Data Num/Class} & single & 213.7 & 212.1 & 233.5 \\
    & multi & 416.2 & 417.8 & 396.4 \\
\bottomrule
\end{tabular}}
\end{center}
\caption{\label{table-result-dataset} Dataset statistics for ACD. \emph{single} denotes the single-aspect sentence and \emph{multi} denotes the multi-aspect one.}
\end{table}

\section{Experiments}
\subsection{Datasets}
\textbf{Aspect Category Detection (ACD)} \; A dataset for few-shot one-class ACD is collected from \emph{YelpAspect} \cite{10.1145/3097983.3098170,li2019exploiting}, which is a large-scale multi-domain dataset for fine-grained sentiment analysis. The 100 aspect categories are split without intersection into 64 classes for training, 16 classes for validation, and 20 classes for testing. 
Table \ref{table-result-dataset} displays the statistics of the dataset. The data in each class is further divided according to the number of the aspects in a sentence, into single-aspect and multi-aspect. Relatively, a multi-aspect example contains more noise when matching sequences. To explore a challenging scenario, the support/query set are both sampled from the multi-aspect set. Fixing this setup, we choose a reference instance as a single- and multi-aspect one.

\vspace{3pt}
\noindent
\textbf{HuffPost} \; It consists of news headlines published in HuffPost between 2012 and 2018 \cite{dataset_huffpost,misra2021sculpting}. \newcite{bao2019few} process the original dataset for few-shot text classification. The number of training, validation, and testing classes are 20, 5, and 16, respectively, where each class has 900 instances. Since the sentences are headlines, they are shorter and less grammatical.

\begin{table*}[t!]
\small
\begin{center}
\setlength{\tabcolsep}{1.0mm}{
\begin{tabular} {l|c|cc|cc|cc|cc}
\toprule
    \multirow{2}{*}{\textbf{Model}} & Use support & \multicolumn{2}{c|}{Match Type} &
    \multicolumn{2}{c|}{ACD \emph{Single}} & \multicolumn{2}{c}{ACD \emph{Multi}} & \multicolumn{2}{c}{HuffPost} \\
    & set of $\mathcal{D}_{test}$ & vector & word & Acc & F1 & Acc & F1 & Acc & F1  \\
    \midrule
    SN & No & \checkmark & & 69.88{\tiny $\pm$1.19} & 69.33{\tiny $\pm$1.16} & 72.12{\tiny $\pm$0.82} & 71.64{\tiny $\pm$0.82} & 62.61{\tiny $\pm$0.55} & 61.87{\tiny $\pm$0.56} \\
    OWP & No & \checkmark & & 72.50{\tiny $\pm$1.22} & 71.96{\tiny $\pm$1.23} & 70.94{\tiny $\pm$0.58} & 70.24{\tiny $\pm$0.65} & 61.72{\tiny $\pm$0.72} & 61.05{\tiny $\pm$0.72} \\
    CA & No & & \checkmark & 79.45{\tiny $\pm$1.17} & 78.97{\tiny $\pm$1.25} & 76.72{\tiny $\pm$0.99} & 76.27{\tiny $\pm$0.96} & 64.02{\tiny $\pm$0.36} & 63.33{\tiny $\pm$0.47}\\
    BiCA & No & & \checkmark & 79.46{\tiny $\pm$0.39} & 79.03{\tiny $\pm$0.46} & 76.81{\tiny $\pm$0.89} & 76.40{\tiny $\pm$0.92} & 64.72{\tiny $\pm$0.77} & 64.20{\tiny $\pm$0.76} \\
    DistilBert & No & & \checkmark & 79.32{\tiny $\pm$1.19} & 78.87{\tiny $\pm$1.36} & 75.16{\tiny $\pm$0.94} & 74.62{\tiny $\pm$0.97} & 64.87{\tiny $\pm$1.32} & 63.91{\tiny $\pm$1.79} \\
    BERT & No & & \checkmark & 79.05{\tiny $\pm$0.98} & 78.61{\tiny $\pm$0.98} & 74.62{\tiny $\pm$0.97} & 74.03{\tiny $\pm$1.03} & 66.02{\tiny $\pm$1.22} & 65.24{\tiny $\pm$1.34} \\
    BERT(p) & No & & \checkmark & 74.99{\tiny $\pm$5.22} & 73.73{\tiny $\pm$6.67} & 76.58{\tiny $\pm$0.90} & 76.07{\tiny $\pm$0.92} & 64.67{\tiny $\pm$0.58} & 63.52{\tiny $\pm$1.26}  \\
    \midrule
    BERT(p)+finetune & Yes & & \checkmark & 83.53{\tiny $\pm$1.11} & 83.30{\tiny $\pm$1.19} & 82.74{\tiny $\pm$0.73} & 82.53{\tiny $\pm$0.75} & 67.43{\tiny $\pm$1.06} & 66.64{\tiny $\pm$1.22} \\
    BERT+funetune & Yes & & \checkmark & 86.10{\tiny $\pm$0.76} & 85.97{\tiny $\pm$0.76} & 82.63{\tiny $\pm$0.77} & 82.43{\tiny $\pm$0.78} & 73.02{\tiny $\pm$0.70} & 72.73{\tiny $\pm$0.69} \\
    BERT+MAML & Yes & & \checkmark & 88.33{\tiny $\pm$2.76} & 88.23{\tiny $\pm$3.07} & 84.99{\tiny $\pm$2.86} & 84.86{\tiny $\pm$3.18} & 73.89{\tiny $\pm$3.28} & 73.66{\tiny $\pm$3.37} \\
    DistilBert+finetune & Yes & & \checkmark & 84.62{\tiny $\pm$1.21} & 84.44{\tiny $\pm$1.26} & 82.15{\tiny $\pm$0.57} & 81.93{\tiny $\pm$0.62} & 69.68{\tiny $\pm$0.83} & 69.22{\tiny $\pm$0.92} \\
    DistilBert+MAML & Yes & & \checkmark & 87.73{\tiny $\pm$0.66} & 87.61{\tiny $\pm$0.67} & 84.93{\tiny $\pm$0.79} & 84.76{\tiny $\pm$0.81} & 72.22{\tiny $\pm$1.60} & 72.00{\tiny $\pm$1.62} \\
    \midrule
    BiCA+finetune & Yes & & \checkmark & 84.62{\tiny $\pm$0.38} & 84.46{\tiny $\pm$0.39} & 82.84{\tiny $\pm$0.97} & 82.70{\tiny $\pm$0.98} & 65.82{\tiny $\pm$0.85} & 65.48{\tiny $\pm$0.89} \\
    BiCA+MAML & Yes & & \checkmark & \bf{89.86$^\dagger${\tiny $\pm$0.65}} & \bf{89.76$^\dagger${\tiny $\pm$0.66}} & \bf{89.80$^\dagger${\tiny $\pm$0.56}} & \bf{89.70$^\dagger${\tiny $\pm$0.57}} & \textbf{74.47$^\ddagger${\tiny $\pm$1.68}} & \textbf{74.20$^\ddagger${\tiny $\pm$1.68}}\\
\bottomrule
\end{tabular}}
\end{center}
\caption{\label{table-result-acd} Experimental results for ACD and HuffPost in terms of accuracy(\%) and macro-f1(\%). We report the average and standard deviation of 5 runs. \emph{Single} indicates that the reference instance is single-aspect. \emph{Multi} indicates setting references as multi-aspect. The marker $^\dagger$ refers to $p$-value$<$0.01 of the T-test compared with DistilBert+MAML. The marker $^\ddagger$ refers to $p$-value$<$0.07 of the T-test compared with DistilBert+MAML.}
\end{table*}

\subsection{Baseline Methods}
\noindent
\textbf{Matching sequences at the vector-level}:

\noindent
{\bf SN} \cite{koch2015siamese} \; Siamese network can capture discriminative features to generalize the predictive power of the network. The input instances are extracted into two vectors, which are compared with \emph{cosine similarity}.

\noindent
{\bf OWP} \cite{kruspe2019one} \; One-way prototypical network designs a 0-vector as the negative prototype. It measures the \emph{Euclidean distance} between an instance with the positive/negative prototypes in the embedding space. 

\vspace{3pt}
\noindent
\textbf{Matching sequences at the word-level}:

\noindent
{\bf CA} \cite{wang2016compare} \; Compare-Aggregate is widely used to match the important units between sequences. It only compares in one direction, i.e., reference-to-candidate.
    
\noindent
{\bf BiCA} \; CA is enhanced into matching sequences bidirectionally (\S\ref{sub:bica}).

\noindent
{\bf DistilBert} \cite{sanh2019distilbert} \; It is a distilled version of BERT (\S\ref{sub:transformer}).

\noindent
{\bf BERT} \cite{devlin-etal-2019-bert} \; It is transformer-based and matching sequences at word-level (\S\ref{sub:transformer}).

\noindent
{\bf BERT(p)} \cite{gao2021making} \; It trains BERT with prompt-based learning. The two sequences are concatenated with ``$\mathtt{? [MASK] ,}$'' like \citeauthor{gao2021making}. The representation of $\mathtt{[MASK]}$ is mapped into word \emph{``yes''} or \emph{``no''}, suggesting that two sequences belong to the same class or not.

\subsubsection{Implementation Details} 
Baseline methods are trained with naive training, which learns the training classes in a meta-task manner, but combines the support set and query set as a whole to optimize parameters. \textbf{+finetune} means that the naive trained models are fine-tuned. \textbf{+MAML} indicates the models are optimized by MAML (Algorithm \ref{alg_optimizing}). During the evaluation, +finetune or +MAML exploit the support set of testing classes in the same way, with the same number of updating steps and learning rate. Thus the only difference between +finetune and +MAML is the parameters are initialized by naive training or MAML training. All testing meta-tasks share the same parameter initialization without mutual interference. The implementation details are described in the Appendix. 

\subsection{Experimental Results}

The experimental results on ACD and HuffPost datasets are displayed in Table \ref{table-result-acd}. The first part in Table \ref{table-result-acd} shows the case that when testing, we only have the reference instance but do not use the support set. By comparing two match types, we find that a finer-granularity matching helps the few-shot one-class scenario gain significantly. This indicates that in the few-shot scenario of text tasks, learning deeper interaction between instances is a better choice. Many previous tasks gain significantly from BERT \cite{wang2020multi} or DistilBert \cite{wright2020transformer}. However, we surprisingly see little performing difference among the five word-level sequence match methods. The transformer-based methods do not have remarkable superiority. A possible reason is that in the unseen classes, it is difficult to discover the \emph{key words/semantics} for matching only given a reference instance. Meanwhile, these models with large-scale parameters may be superior in the data-driven tasks \cite{gururangan-etal-2020-dont}.

Additionally, though the scale of the support set is small, exploiting it by fine-tuning or MAML can bring significant improvements. This also explains our previous guess that the support set will provide \emph{key words/semantics} to match. Meanwhile, on sequence match methods, including BiCA, BERT and DistilBert, MAML outperforms fine-tuning in most situations. This indicates the importance of a good parameter initialization not only for small models but also for large pre-trained models in a few-shot problem.

It is also found that prompt-based fine-tuning, i.e. BERT(p), is also less-performed than BiCA+MAML. The possible reasons are: first, the objective of one-class sequence match is not consistent with the pre-training of language models. Thus the knowledge of transformer models might not be fully leveraged; 
second, prompt-based fine-tuning may achieve better results by other huge-scale pre-trained models, such as RoBERTa-large, GPT-3 \cite{gao2021making}.


Finally, it is worth noting that BiCA+MAML consistently outperforms BERT+MAML and DistilBert+MAML. Compared with transformer models, BiCA+MAML has fewer parameters, suggesting that the classical methods are still worth revisiting in the large pre-trained models' dominant era. We further see another interesting phenomenon. BiCA gains significantly from MAML but slightly improves by using fine-tuning. Contrarily, the transformer-based method gains much from fine-tuning. It is possible that the pre-trained BERT already contains abundant knowledge, suggesting a good initialization for fine-tuning. Meanwhile, BiCA is a classical model with much fewer parameters, which is easier for MAML to learn a good initialization. Hence, MAML has a more significant contribution to BiCA.

\subsection{Discussion}
\label{subsec:discussion}
In this section, an empirical comparison, between two kinds of sequence match approaches, including classical and transformer-based ones, is presented. Because DistilBert+MAML and BERT+MAML are comparable, as shown in Table \ref{table-result-acd}. Meanwhile, the parameter scale of DistilBert is smaller, which is chosen in the following study. We randomly sample 12 batches from the testing classes, and obtain the extracted features of the query set. In each batch, we have 5 meta-tasks, each of them has 10 support instances and 10 query instances. Thus, the total number of features is $5\times{10}\times{12}=600$. The features are visualized by t-SNE \cite{maaten2008visualizing}. 

\begin{figure*}[t]
\centering
\includegraphics[width=0.98\textwidth]{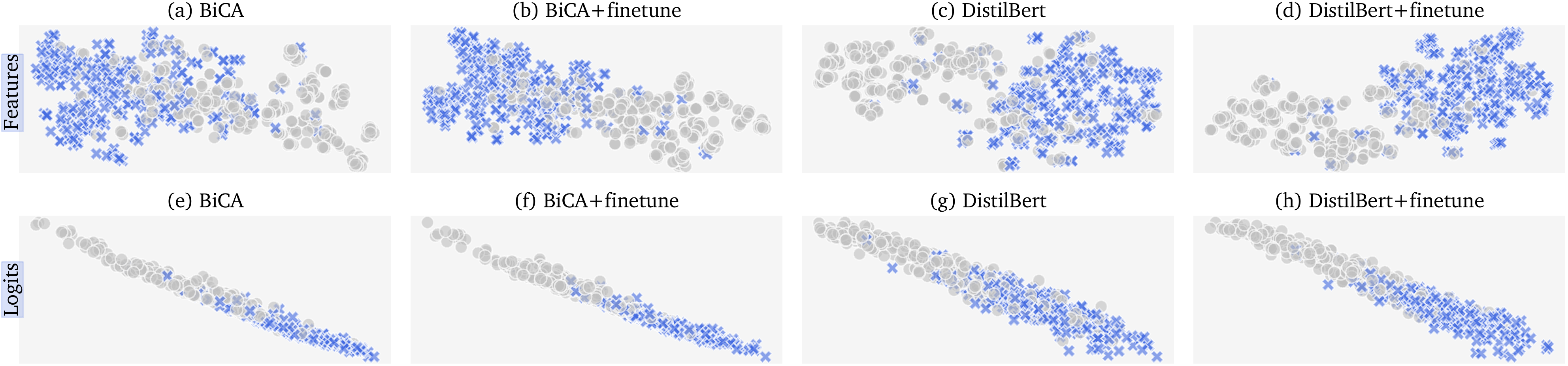}
\caption{Effects of fine-tuning on BiCA and DistilBert for ACD, where the reference instance is single-aspect. For a fair comparison, all models only have one linear output layer. In the top row, we depict PCA plots of the features before the output layer. The feature dimension in BiCA is 500 and which in DistilBert is 768. In the bottom row, we directly plot the 2-dimensional logits output.}
\label{fig:bica_single_fine-tuning}
\end{figure*}

\begin{figure*}[t]
\centering
\includegraphics[width=0.98\textwidth]{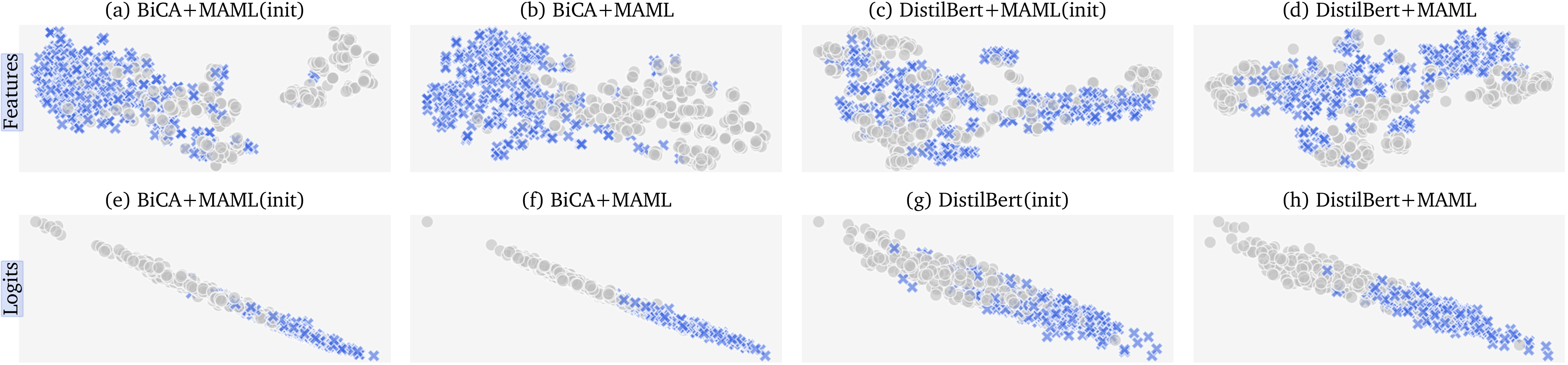}
\caption{Effects of MAML on BiCA and DistilBert for ACD, where the reference instance is single-aspect. ``(init)'' indicates that the model learned by MAML training directly predicts the query set of testing meta-tasks without exploiting the support set.}
\label{fig:bica_single_maml}
\end{figure*}

\subsubsection{Comparison of Features}
We compare the effects of MAML with simple fine-tuning in Figure \ref{fig:bica_single_fine-tuning} and Figure \ref{fig:bica_single_maml}, respectively. In Figure \ref{fig:bica_single_fine-tuning}, it can be seen that fine-tuning can help the features learned by BiCA and DistilBert both become more separable (plot a-d). This is also reflected by the 2-dimensional logits output (e-g). 

In Figure \ref{fig:bica_single_maml}, it can be observed that in MAML training, exploiting the support set can also make the features more discriminative in BiCA (a-b), and so does the logits output (e-f). The features (plot b in Figure \ref{fig:bica_single_fine-tuning} and plot b in Figure \ref{fig:bica_single_maml}) validates that MAML is more effective than fine-tuning in BiCA.

\begin{table}[t!]
\small
\begin{center}
\begin{tabular} {l|ccc}
\toprule
    \multirow{2}{*}{Model} & \multicolumn{2}{c}{ACD} & \multirow{2}{*}{HuffPost} \\
    & \emph{Single} & \emph{Multi} & \\
    \midrule
    BiCA & 1.74e-3 & 1.23e-3 & 2.66e-4 \\
    BiCA+finetune & 1.77e-3 & 1.22e-3 & 2.84e-4 \\
    BiCA+MAML(init) & 1.79e-3 & 1.67e-3 & 3.53e-4 \\
    BiCA+MAML & \bf{1.92e-3} & \bf{2.09e-3} & \bf{5.85e-4} \\
    \midrule
    DistilBert & 3.35e-3 & 2.70e-3 & 1.51e-3 \\
    DistilBert+finetune & 4.01e-3 & 3.39e-3 & 1.78e-3 \\
    DistilBert+MAML(init) & 8.57e-3 & 1.15e-2 & 7.40e-3 \\
    DistilBert+MAML & \bf{8.98e-3} & \bf{1.32e-2} & \bf{7.76e-3} \\
    \midrule
    BERT & 3.89e-3 & 3.91e-3 & 4.62e-3 \\
    BERT+finetune & 4.83e-3 & 5.56e-3 & 6.43e-3 \\
    BERT+MAML(init) & 2.14e-1 & 1.71e-1 & {\bf 2.63e-1}  \\
    BERT+MAML & \bf{2.21e-1} & \bf{1.99e-1} & 2.58e-1 \\
    
\bottomrule
\end{tabular}
\end{center}
\caption{\label{table-result-cov} $Cov\_Score$ of various models. The largest is marked in bold for each sequence match model.}
\end{table}

Interestingly, the phenomenon of DistilBert+MAML is completely different. It is found that the features show less separability (c-d), while the logits output is well distinguished (g-h). This indicates the features are linearly separable in high dimensions, i.e. 768. Recalling the purpose of PCA (Principal Component Analysis) \cite{abdi2010principal}, it defines an orthogonal linear transformation that transforms the data into a new coordinate system and preserves the greatest variance. We guess that the unique phenomena (c-d) are caused by less-orthogonal feature dimensions. To verify this assumption, we compute the covariance matrix based on extracted features in various models. Each element in the matrix indicates the correlation between two dimensions in feature, where a larger score means a higher correlation. We define the $Cov\_Score$ as below, which is the average absolute value of the covariance matrix.
\begin{equation}
    Cov\_Score = \mathrm{avg}|\mathrm{Cov}(\mathbf{F}-\mathrm{rowavg}(\mathbf{F}))|
    \label{eq:cov}
\end{equation}
where $\mathbf{F}$ is the extracted features.

\begin{figure}[t]
\centering
\includegraphics[width=0.48\textwidth]{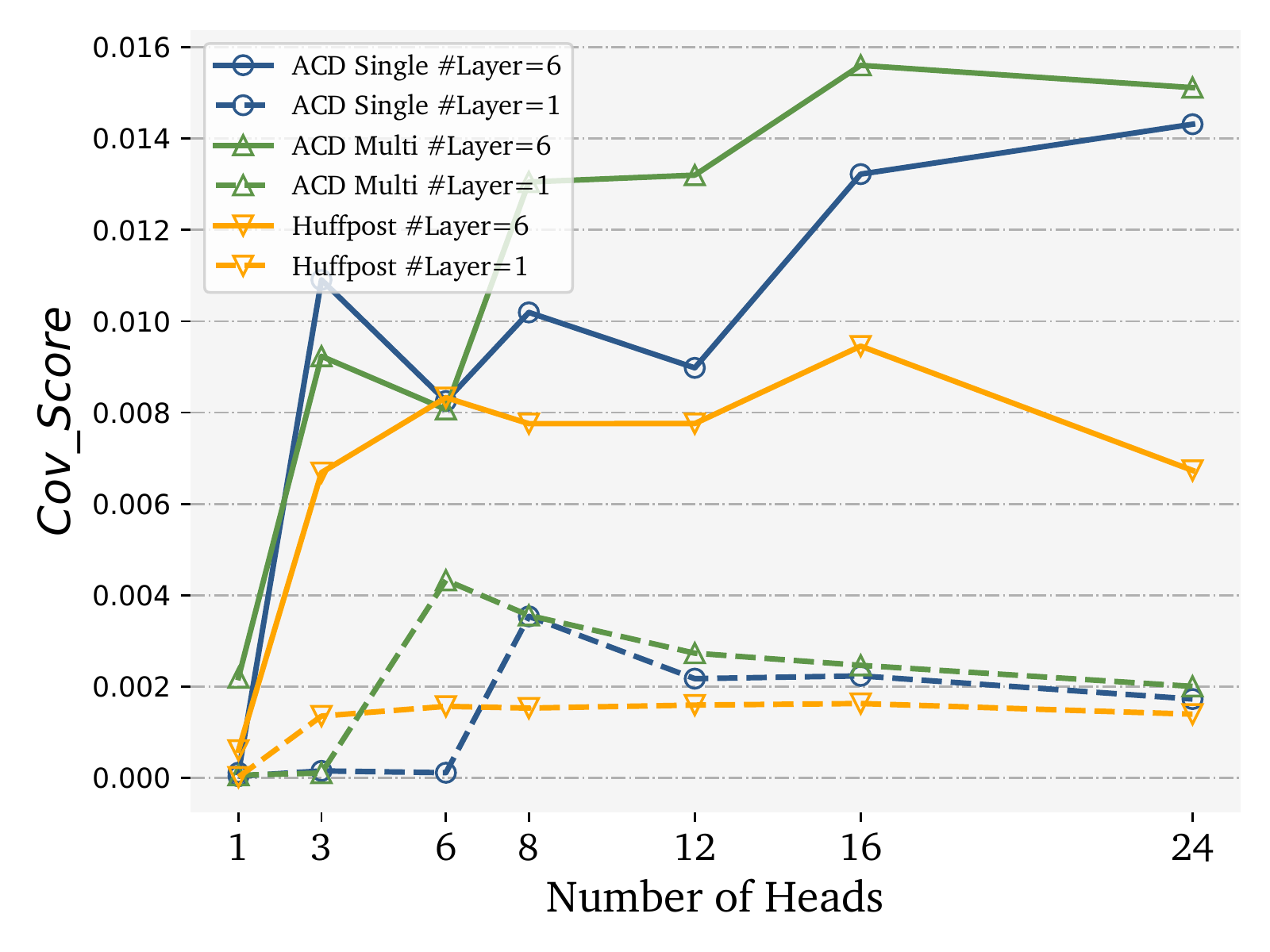}
\caption{$Cov\_Score$ of the features learned by DistilBert+MAML, setting different numbers of layers, and numbers of heads in self-attention.}
\label{fig:cov_layer_head}
\end{figure}

In Table \ref{table-result-cov}, the $Cov\_Score$ of three sequence match approaches are presented separately. Firstly, compared with BiCA+finetune, BiCA+MAML extracts features with slightly higher $Cov\_Score$. However, for DistilBert and BERT, MAML dramatically increases the score, which is more significant in BERT. As the scores indicate, DistilBert+MAML really extracts features with less-orthogonal dimensions. A possible reason is that MAML trains the model in a bi-level manner (see Algorithm \ref{alg_optimizing}). DistilBert has deeper layers and large-scale parameters. The gradients are propagated through deeper layers, causing the feature extraction learned insufficiently while the last linear layer is trained adequately. Thus the logits are separable while features are not, as plots (c, d, g, h) shown in Figure \ref{fig:bica_single_maml}. The above phenomenon become also serious in BERT+MAML, because BERT has more layers compared with DistilBert, leading to larger $Cov\_Score$. 

The plots of other datasets are displayed in the Appendix. The empirical observations are similar.

\subsubsection{DistilBert+MAML: Layers and Heads?}
To further investigate why the features learned by DistilBert+MAML have high-correlation dimensions, we plot the $Cov\_Score$ of multiple variants of DistilBert+MAML in Figure \ref{fig:cov_layer_head}. The horizontal ordinate indicates the number of heads in self-attention, which should divide 768 exactly (768 is the dimension size of the hidden states in DistilBert and BERT models). 

It is first seen that the score shows an increasing trend as the number of heads grows. Meanwhile, by comparing the curves between \#Layer=1 and \#Layer=6, we observe that the $Cov\_Score$ also becomes larger as the layers of DistilBert increase. We draw an empirical conclusion that the high-correlation of feature dimensions in DistilBert+MAML is caused by the multiple layers and heads of DistilBert.   

\begin{figure}[t]
\centering
\includegraphics[width=0.48\textwidth]{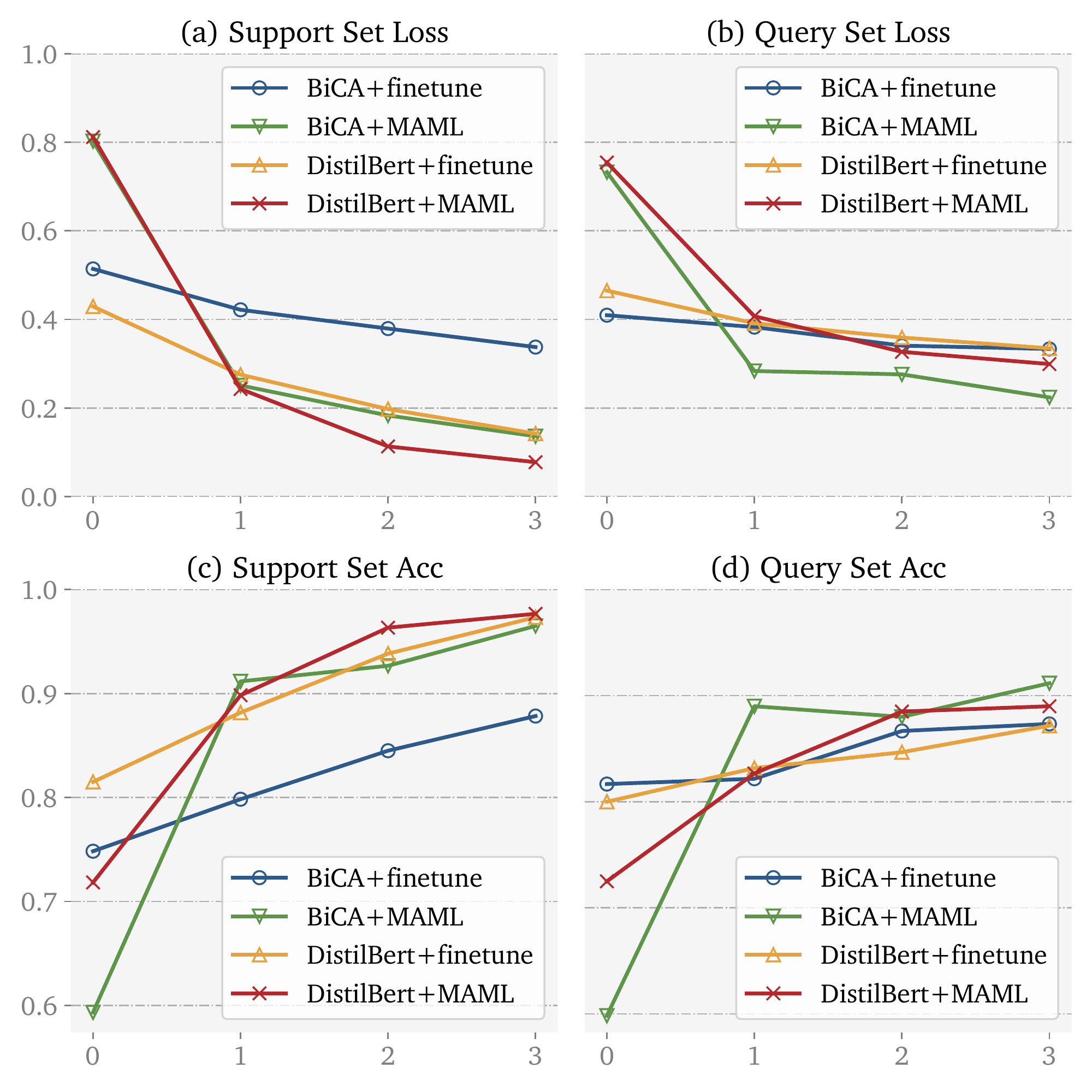}
\caption{Loss and accuracy curves in 3 updating steps.}
\label{fig:bica_single_update_steps}
\end{figure}

\subsubsection{Initialization for Loss Sensitivity or a Good Performance Start}
In Figure \ref{fig:bica_single_update_steps}, we depict the average batch loss and accuracy in the 3 update steps for ACD with single-aspect references. Step 0 indicates the model is trained by naive training or MAML without using the testing support set.

Concretely, it can be seen that for both BiCA and DistilBert, MAML leads to faster loss degradation than fine-tuning (plot a). The loss in MAML declines to the bottom even within one updating step. We also observe a rapid accuracy increase (c). Overall, MAML outperforms fine-tuning for both BiCA and DistilBert. However, is MAML always a good choice? The answer is negative. In step 0 (b and d), we see a good parameter initialization does not lead to a good performance start. BiCA and DistilBert learned by naive training achieve lower losses and better accuracy scores without updating parameters. A possible explanation is that they have different training objectives. Naive training aims to facilitate task-agnostic sequence matching. While the bi-level optimization in MAML focuses on promoting the generalization ability of models after seeing a few support examples. 

\subsubsection{Various Numbers of Support Instances}

\begin{figure}[t]
\centering
\includegraphics[width=0.48\textwidth]{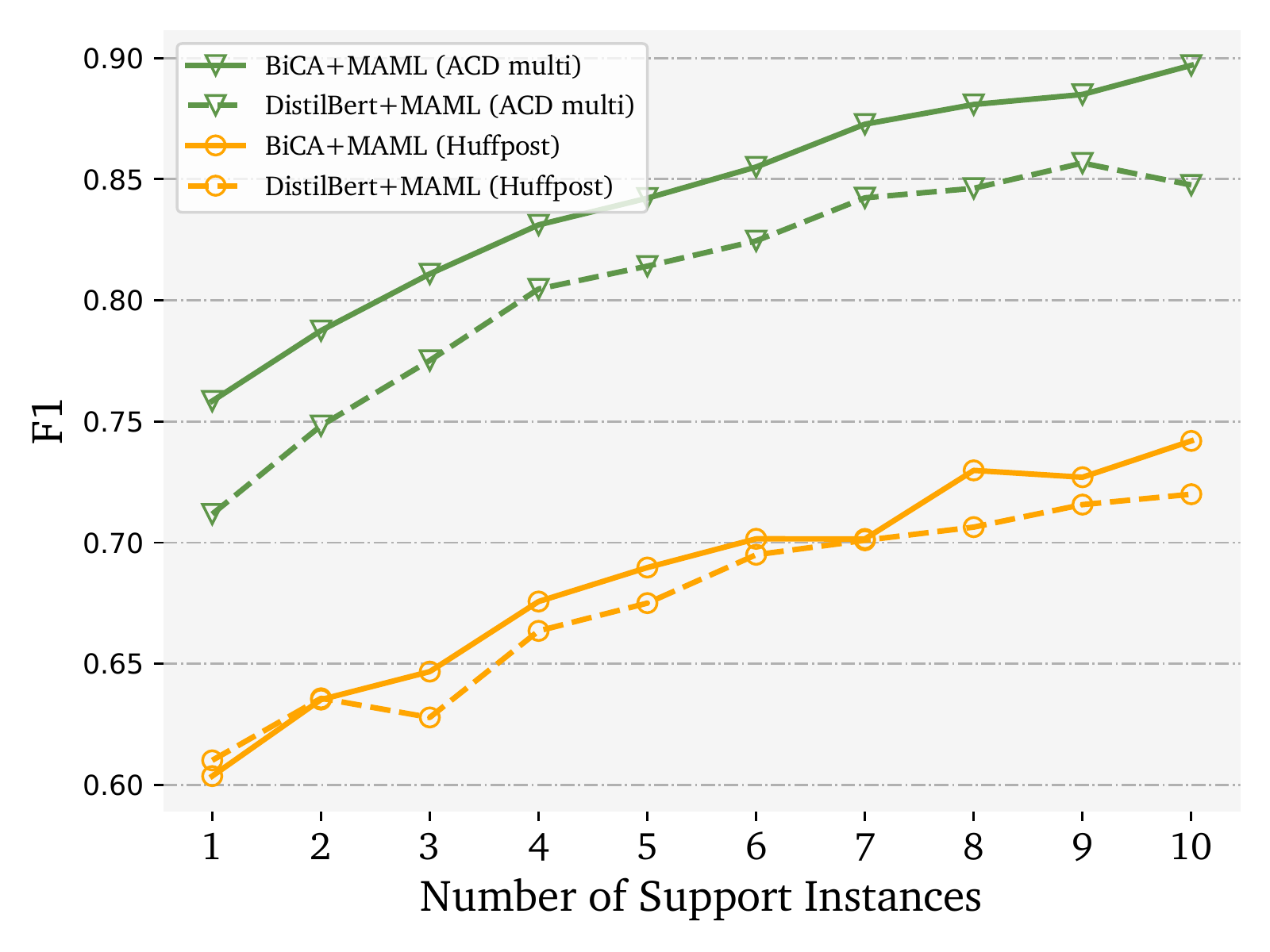}
\caption{F1 scores of BiCA+MAML and DistilBert+MAML by setting different numbers of support instances.}
\label{fig:num_support}
\end{figure}

We further explore the performances of BiCA+MAML and DistilBert+MAML under various support set scales. As displayed in Figure \ref{fig:num_support}, the number of support instances ranges from 1 to 10. For a fair comparison, the number of query instances is all set to 10. Firstly, it can be seen that as the support set scale grows, the performances on the query set present an increasing trend. Secondly, we also observe that BiCA+MAML outperforms DistilBert+MAML in most experimental settings. This indicates that classical sequence match is a competitive few-shot one-class learner.

\section{Conclusion}
In this work, we revisit the classical sequence match approaches and find that with meta-learning, the classical method can significantly outperform transformer models in the few-shot one-class scenario. The training cost is greatly reduced. Furthermore, an empirical study is made to explore the effects of simple fine-tuning and meta-learning. Interestingly, although meta-learning is more effective than simple fine-tuning on both sequence match approaches, it makes the transformer features have high correlation dimensions. The correlation is closely related to the number of layers and heads in the transformer models. We hope this work could provide insights for future research on few-shot problems and transformer models.

\section*{Acknowledgements}
We sincerely thank all the anonymous reviewers for providing valuable feedback. This work is supported by the key program of National Science Fund of Tianjin, China (Grant No. 21JCZDJC00130) and the Basic Scientific Research Fund, China (Grant No. 63221028).

\bibliography{custom}
\bibliographystyle{acl_natbib}

\clearpage
\appendix
\begin{figure*}[t]
\centering
\includegraphics[width=0.98\textwidth]{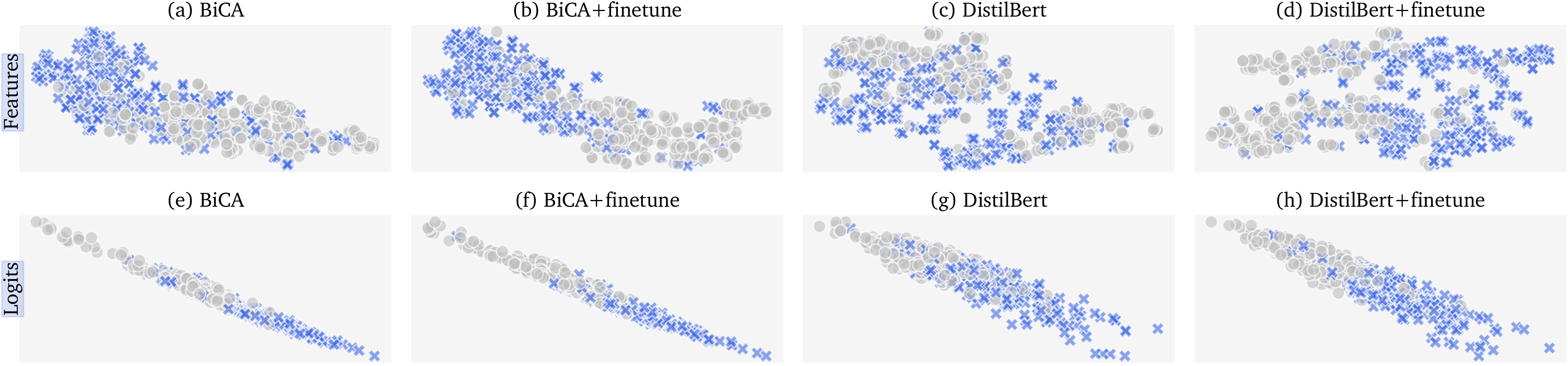}
\caption{Effects of fine-tuning on aspect category detection, where the reference instance is multi-aspect.}
\label{fig:acd_multi_fine-tuning}
\end{figure*}

\begin{figure*}[t]
\centering
\includegraphics[width=0.98\textwidth]{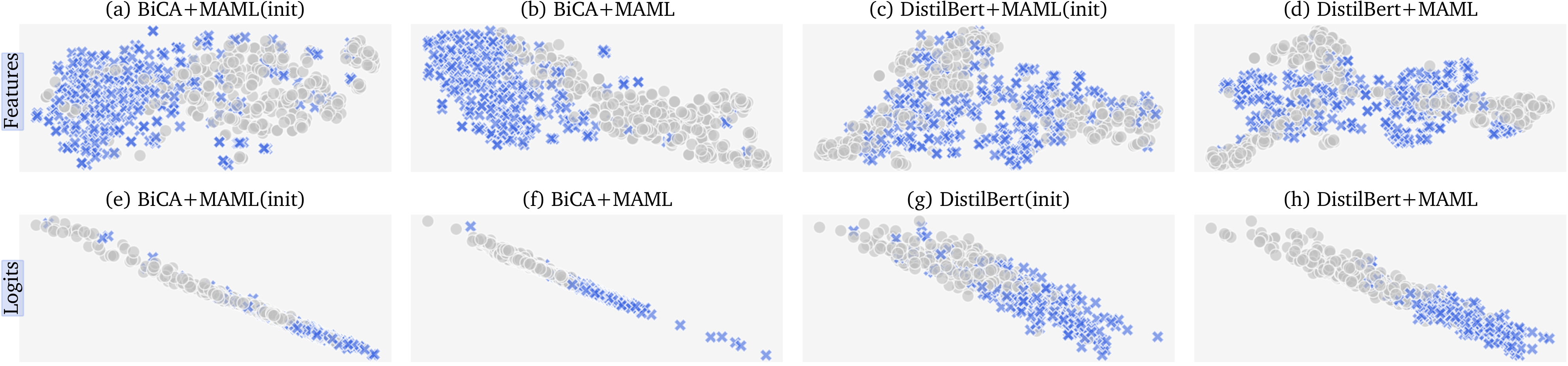}
\caption{Effects of MAML on aspect category detection, where the reference instance is multi-aspect.}
\label{fig:acd_multi_maml}
\end{figure*}

\begin{figure*}[t]
\centering
\includegraphics[width=0.98\textwidth]{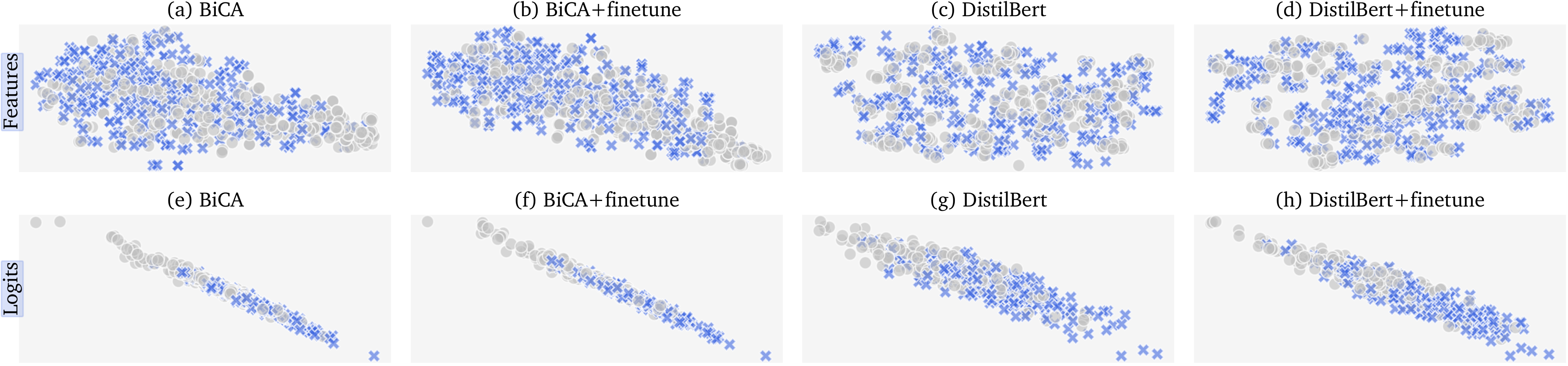}
\caption{Effects of fine-tuning on HuffPost.}
\label{fig:huffpost_fine-tuning}
\end{figure*}

\begin{figure*}[t]
\centering
\includegraphics[width=0.98\textwidth]{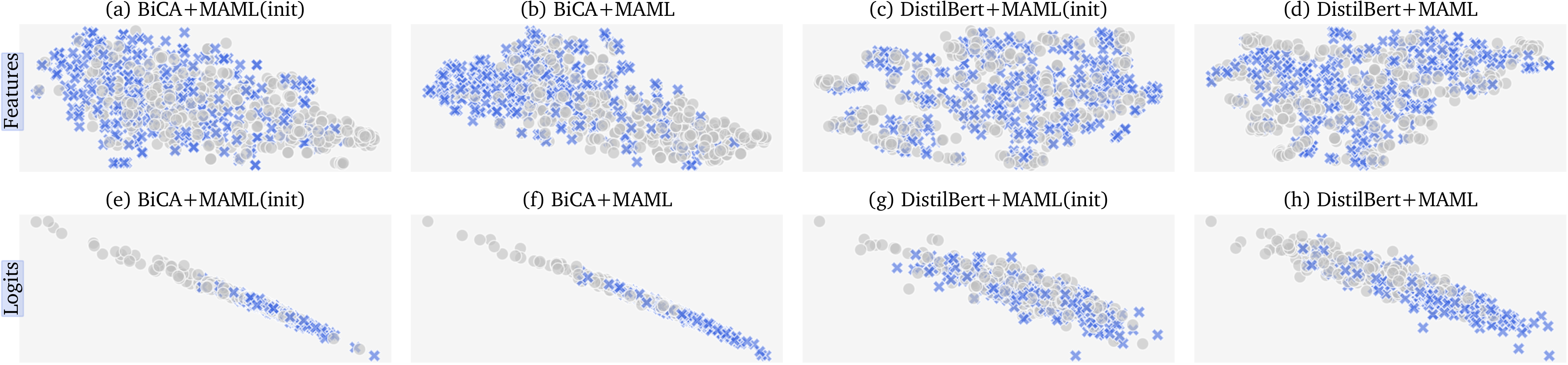}
\caption{Effects of MAML on HuffPost.}
\label{fig:huffpost_maml}
\end{figure*}

\section{Additional Experimental Results}
The visualizations of aspect category detection with multi-aspect reference instance are displayed in Figure \ref{fig:acd_multi_fine-tuning} and Figure \ref{fig:acd_multi_maml}. 

For HuffPost dataset, the visualizations are displayed in Figure \ref{fig:huffpost_fine-tuning} and \ref{fig:huffpost_maml}.

\section{Reproducibility}
\subsection{Computing Infrastructure}
All experiments are conducted on the same hardware and software. We use a single NVIDIA A6000 GPU with 48GB of RAM.

\subsection{Average Running time}
The average running time of each model is shown in Table \ref{table-runing-time}.

\subsection{Number of Parameters}
The number of parameters of each model is shown in Table \ref{table-parameter}.

\subsection{Datasets}
The datasets are available at \url{https://github.com/hmt2014/FewOne}.

\subsection{Implementation Details}
\subsubsection{Classical Methods}
All baselines and our model are implemented by Pytorch. We initialize word embeddings with 50-dimension GloVE vectors \cite{pennington2014glove}. In the aggregation module, the channels of the CNNs for input and output are both 50. The kernel sizes of five CNNs are [1, 2, 3, 4, 5], respectively. $\mathrm{Relu}$ is the activation function for CNN. We adopt a dropout of 0.1 after both the comparison and CNN in aggregation. $\mathrm{MLP}$ is a single linear layer.

The batch size is $|\mathcal{C}_p|=5$, indicating a batch comprises 5 meta-tasks. The instance number in the support set $|S|$ and query set $|Q|$ are both set to 10. Every epoch we randomly sample 400 batches for training, 300 batches for validation and 300 batches for testing. The average results of the testing batches are reported. We exploit an early stop strategy during training if the macro-f1 score on the validation set does not improve in 3 epochs, and the best model is chosen for evaluation. 

We describe the training details as the format of (optimizer, learning rate, other information):

\vspace{3pt}
\noindent
\textbf{Naive Training}: Adam, 1e-3, early stop. 

\vspace{3pt}
\noindent
\textbf{+finetune} \; SGD, 0.1, 3 updating steps.

\vspace{3pt}
\noindent
\textbf{+MAML} \; 

The inner-level: $\alpha$=0.1. 

The outer-level: Adam, $\beta$=1e-3.

When testing: SGD, 0.1, 3 updating steps.

\subsubsection{Transformer-based Methods}
The output hidden state of $\mathtt{[CLS]}$ in DistilBert (distilbert-base-uncased) and BERT (bert-base-uncased) is exploited for classification.

\vspace{3pt}
\noindent
\textbf{Naive Training} \; Adam, 2e-5, 5 epochs.

\vspace{3pt}
\noindent
\textbf{+finetune} \; Adam, 2e-5, 3 updating steps.

\vspace{3pt}
\noindent
\textbf{+MAML} \; 

The inner level: $\alpha$=2e-3. 

The outer-level: Adam, $\beta$=2e-5. 

When testing: Adam, 2e-5, 3 updating steps.

We use early stop for DistilBert+MAML since the model does not learn optimally within 5 epochs.

\begin{table}[t!]
\small
\begin{center}
\setlength{\tabcolsep}{2.4mm}{
\begin{tabular} {l|ll}
\toprule
    Method &  ACD & Huffpost \\
    \midrule
    SN & 1,437,950 & 1,393,050 \\
    OWP & 1,437,950 & 1,393,050 \\
    CA & 1,476,202 & 1,431,302 \\
    \midrule
    BiCA  & \multirow{3}{*}{1,476,702} & \multirow{3}{*}{1,431,802} \\
    +finetune  \\
    +MAML \\
    \midrule
    DistilBert  & \multirow{3}{*}{66,364,418} & \multirow{3}{*}{66,364,418}  \\
    +finetune \\
    +MAML \\
    \midrule
    BERT & \multirow{3}{*}{109,483,778} & \multirow{2}{*}{109,483,778} \\
    +finetune \\
    +MAML \\
    \midrule
    BERT(p) & \multirow{2}{*}{133,545,786} & \multirow{2}{*}{133,545,786} \\
    +finetune \\
\bottomrule
\end{tabular}}
\end{center}
\caption{\label{table-parameter} Number of parameters in each model.}
\end{table}

\begin{table}[t!]
\small
\begin{center}
\setlength{\tabcolsep}{1.2mm}{
\begin{tabular} {l|lll}
\toprule
    \multirow{2}{*}{Method} &  \multicolumn{2}{c}{ACD} & \multirow{2}{*}{Huffpost} \\
    & single & multi & \\
    \midrule
    SN & 6m47s & 11m44s & 20m5s \\
    OWP & 13m57s & 17m49s & 11m45s \\
    CA & 21m36s & 35m17s & 18m2s \\
    \midrule
    BiCA  & 22m26s & 18m5s & 11m11s \\
    +finetune & 20s & 20s & 23s  \\
    \midrule
    BiCA+MAML & 39m8s & 1h11m57s & 40m53s \\
    \midrule
    DistilBert  & 21m2s & 21m26s & 21m56s  \\
    +finetune & 4m45s & 4m43s & 4m45s \\
    \midrule
    DistilBert+MAML & 4h32m2s & 4h3m38s & 4h46m4s \\
    \midrule
    BERT & 39m33s & 40m18s & 41m27s\\
    +finetune & 8m40s & 8m41s & 9m14s \\
\bottomrule
\end{tabular}}
\end{center}
\caption{\label{table-runing-time} Average runing time of each model.}
\end{table}

\end{document}